\documentclass[10pt,twocolumn,letterpaper]{article}

\usepackage[pagenumbers]{cvpr}            %

\usepackage{graphicx}
\usepackage{amsmath}
\usepackage{amssymb}
\usepackage{booktabs}
\usepackage{mathtools}
\usepackage{amsthm} 
\usepackage{bm}
\usepackage{subcaption,booktabs} %
\usepackage{array}
\usepackage{import}
\usepackage{example}

\usepackage[accsupp]{axessibility}

\newcolumntype{C}[1]{>{\centering\arraybackslash}m{#1}}
\newcolumntype{R}[1]{>{\raggedleft\arraybackslash}m{#1}}
\newcolumntype{P}[1]{>{\raggedright\arraybackslash}p{#1}}
\newcolumntype{M}[1]{>{\centering\arraybackslash}m{#1}}
\usepackage{multirow}
\usepackage{makecell}
\usepackage[table,dvipsnames]{xcolor}

\newcommand{\wj}[1]{{\color{black}#1}}

\newcommand{\tu}[1]{{\color{black}#1}}

\newcommand{\camera}[1]{{\color{black}#1}}
\usepackage[pagebackref,breaklinks,colorlinks]{hyperref}

\usepackage[capitalize]{cleveref}
\crefname{section}{Sec.}{Secs.}
\Crefname{section}{Section}{Sections}
\Crefname{table}{Table}{Tables}
\crefname{table}{Tab.}{Tabs.}

\begin{document}

\title{A Bag-of-Prototypes Representation for Dataset-Level Applications}

\author{Weijie Tu$^{1 }$ \quad Weijian Deng$^{1 }$ \quad Tom Gedeon$^{2 }$ \quad Liang Zheng$^{1 }$ \\
$^{1}$Australian National University \quad \ $^{2}$
Curtin University}
\maketitle

\begin{abstract}
This work investigates dataset vectorization for two dataset-level tasks: assessing training set suitability and test set difficulty. The former measures how suitable a training set is for a target domain, while the latter studies how challenging a test set is for a learned model. Central to the two tasks is measuring the underlying relationship between datasets. This needs a desirable dataset vectorization scheme, which should preserve as much discriminative dataset information as possible so that the distance between the resulting dataset vectors can reflect dataset-to-dataset similarity. To this end, we propose a bag-of-prototypes (BoP) dataset representation that extends the image-level bag consisting of patch descriptors to dataset-level bag consisting of semantic prototypes. Specifically, we develop a codebook consisting of $K$ prototypes clustered from a reference dataset. Given a dataset to be encoded, we quantize each of its image features to a certain prototype in the codebook and obtain a $K$-dimensional histogram. Without assuming access to dataset labels, the BoP representation provides a rich characterization of the dataset semantic distribution. 
Furthermore, BoP representations cooperate well with Jensen-Shannon divergence for measuring dataset-to-dataset similarity. Although very simple, BoP consistently shows its advantage over existing representations on a series of benchmarks for two dataset-level tasks.
\end{abstract}

\section{Introduction}
\label{sec:intro}

Datasets are fundamental in machine learning research, forming the basis of model training and testing~\cite{dataset_survey,deng2009imagenet, lin2014microsoft,liang2022advances}. While large-scale datasets bring opportunities in algorithm design, there lack proper tools to analyze and make the best use of them \cite{mazumder2022dataperf,liang2022advances,aroyo2022data}. %
Therefore, as opposed to traditional algorithm-centric research where improving models is of primary interest, the community has seen a growing interest in understanding and analyzing the data used for developing models \cite{liang2022advances,mazumder2022dataperf}.
Recent examples of such goal include data synthesis \cite{fu2022stylegan}, data sculpting \cite{eyuboglu2022dcbench,liang2022advances}, and data valuation \cite{mazumder2022dataperf,aroyo2022data,ghorbani2019data}. 
These tasks typically focus on {individual} sample of a dataset. In this work, we aim to understand nature of datasets from a dataset-level perspective.

This work considers two dataset-level tasks: suitability in training and difficulty in testing.
\textbf{First}, training set suitability denotes whether a training set is suitable for training models for a target dataset. In real-world applications, we are often provided with multiple training sets from various data distributions (\eg, universities and hospitals). Due to distribution shift, their trained models have different performance on the target dataset.
Then, it is of high practical value to select the most suitable training set for the target dataset.
\textbf{Second}, test set difficulty means how challenging a test set is for a learned model. In practice, test sets are usually unlabeled and often come from different distributions than that of the training set.
Measuring the test set difficulty for a learned model helps us understand the model reliability, thereby ensuring safe model deployment.

The core of the two dataset-level tasks is to measure the relationship between datasets. For example, 
\wj{a training set is more suitable for learning a model if it is more similar to the target dataset.}
To this end, we propose a vectorization scheme to represent a dataset. Then, the relationship between a pair of datasets can be simply reflected by the distance between their representations.
Yet, it is challenging to encode a dataset as a representative vector, because \textit{(i)} a dataset has a different cardinality (number of images) and \textit{(ii)} each image has its own semantic content (\eg, category).
It is thus critical to find an effective way to aggregate all image features to uncover dataset semantic distributions.

In the literature, some researchers use the first few moments of distributions such as feature mean and co-variance to represent datasets \cite{sun2016deep, sun2016return, peng2019moment, deng2021labels,vanschoren2018meta}. While being computational friendly, these methods do not offer sufficiently strong descriptive ability of a dataset, such as class distributions, and thus have limited effectiveness in \wj{assessing attributes related to semantics}. 
There are also some methods learn \textit{task-specific} dataset representations \cite{peng2020domain2vec, achille2019task2vec}. For example, given a dataset with labels and a task loss function, Task2Vec \cite{achille2019task2vec} computes an embedding based on estimates of the Fisher information matrix associated with a probe network's parameters. 
While these task-specific representations are able to predict task similarities, they are not suitable for characterizing \wj{dataset properties of interest.}
They require training a network on the specific task \cite{achille2019task2vec} or on multiple datasets \cite{peng2020domain2vec}, so they are not effective in assessing the training set suitability. Additionally, they require image labels for the specific task, so they cannot be used to measure the difficulty of unlabeled test sets.

In this work, we propose a simple and effective bag-of-prototypes (BoP) dataset representation. %
Its computation starts with partitioning the image feature space into semantic regions through clustering, where the region centers, or prototypes, form a codebook. Given a new dataset, we quantize its features to their corresponding prototypes and compute an assignment histogram, which, after normalization, gives the BoP representation. The dimensionality of BoP equals the codebook size, which is usually a few hundred and is considered memory-efficient. Meanwhile, the histogram computed on the prototypes is descriptive of the dataset semantic distribution. 

Apart from being low dimensional and semantically rich, BoP has a few other advantages. \textbf{First}, while recent works in task-specific dataset representation usually require full image annotations and additional learning procedure \cite{peng2020domain2vec, achille2019task2vec}, the computation of BoP does not rely on any. It is relatively efficient and allows for \wj{unsupervised assessment of dataset attributes.} %
\textbf{Second}, BoP supports dataset-to-dataset similarity measurement through Jensen-Shannon divergence. We show in our experiment that this similarity is superior to commonly used metrics such as Fr\'echet distance \cite{frechet1957distance} and maximum mean discrepancy \cite{gretton2006kernel} in two dataset-level tasks.

\section{Related Work}

\textbf{{Dataset representations.}} A common practice is to use simple and generic statistics as dataset representations \cite{sun2016deep, sun2016return, peng2019moment, deng2021labels,vanschoren2018meta}. For example,
Peng \etal~\cite{peng2019moment} use the first moment to represent a dataset. 
Deng \etal~\cite{deng2021labels} use global feature mean and co-variance as dataset representations. 
Vanschoren \etal~\cite{vanschoren2018meta} find dataset cardinality (the number of images/classes) useful to encode a dataset. 
These methods have limited descriptive ability, whereas BoP is more semantically descriptive. 
Moreover, it is feasible to learn a \textit{task-specific} dataset representation~\cite{peng2020domain2vec, achille2019task2vec,ying2018transfer, zhong2019ghostvlad}. 
For example, 
Ying~\etal~\cite{ying2018transfer} learn transfer skills from previous transfer learning experiences for future target tasks.
Achille~\etal~\cite{achille2019task2vec} propose to learn a task embedding based on the estimate of Fisher information matrix associated with a task loss. 
Compared with these task-specific representations, BoP is hand-crafted, avoiding computation overheads incurred by end-to-end learning. It is thus efficient in measuring training set suitability without training any models. Moreover, BoP require no image labels,
making it more suitable for assessing the difficulty of unlabeled test sets.

\textbf{{Dataset-to-dataset similarity.}} We briefly review three strategies. \textbf{First}, some dataset similarity measures are developed in the context of domain adaptation \cite{ben2006analysis, shai2010hdh, acuna2021f, zhang2019bridging}.
They typically depend on a loss function and hypothesis class, and use a supremum of that function class to quantify the similarity of datasets. %
(\eg, $\mathcal{H}\Delta\mathcal{H}$-divergence \cite{shai2010hdh}, $f$-divergence \cite{acuna2021f}, and $\mathcal{A}$-distance \cite{ben2006analysis}). 
\textbf{Second}, dataset distance can be computed based on optimal transport \cite{alvarez2020geometric, tan2021otce, delon2020wasserstein}. For example, %
{the squared Wasserstein metric Fr\'echet distance \cite{frechet1957distance} is widely used in comparing the distribution discrepancy of generated images with the distribution of real images \cite{heusel2017gans}.}
To better leverage the geometric relationship between datasets, Alvarez~\etal~\cite{alvarez2020geometric} use labels to guide optimal transport towards class-coherent matches.
\textbf{Third}, existing dataset representations can be used to compute dataset distance \cite{gretton2006kernel,peng2019moment,sun2016deep,tzeng2014deep}. 
For example, 
maximum mean discrepancy (MMD) \cite{gretton2006kernel} computes the distance between mean elements of distributions on the probability space.
Peng \etal~\cite{peng2019moment} eliminate dataset discrepancy by matching datasets moments.
CORAL \cite{sun2016deep} uses the second-order statistics of datasets to measure distance.
This work is in the third category and uses JS divergence between BoP representations to calculate dataset-to-dataset similarity.

\textbf{Assessment of training dataset suitability.} 
Recent works have focused on understanding the importance of \textit{individual training instances} in training of neural networks \cite{mazumder2022dataperf,aroyo2022data,ghorbani2019data,jiang2020characterizing}. For example, Data Shapley \cite{ghorbani2019data} and Consistency Score \cite{jiang2020characterizing} are proposed to evaluate the value of each data instance. 
Some methods identify ``difficult" instances based on the information of training dynamics \cite{toneva2018empirical,swayamdipta2020dataset,baldock2021deep}. 

Different from the above approaches, this work studies the suitability of an \textit{entire} training set. Given multiple training datasets from different data distributions, the focus is to choose the most appropriate training dataset for the target domain.
{Dataset-to-dataset similarity can be used for this goal.}
Intuitively, if a training dataset has high similarity with a target dataset, the model trained on it is expected to be more performant and vice versa. 
In this work, we use BoP representation coupled with simple JS divergence to calculate dataset-to-dataset similarity and demonstrate its effectiveness in accessing training set suitability.

\textbf{{Assessing test set difficulty without ground truths.}} The goal of this task (also known as unsupervised accuracy estimation) is to predict the accuracy of a given model on various unlabeled test sets.
Existing methods usually use a representation of the test set for accuracy prediction \cite{deng2021labels,guillory2021predicting,Deng:ICML2021,garg2022leveraging,chen2021detecting}.
Normally this representation is derived from classifier outputs, such as image features \cite{deng2021labels}, prediction logits \cite{garg2022leveraging}, average softmax scores \cite{guillory2021predicting}. Then, regression is used to establish the relationship between this representation and model test accuracy under various testing environments. 
Compared with existing dataset features, the BoP representation better characterizes the semantic distribution of training and test sets and thus can be effectively used for model accuracy prediction. 

\section{Methodology}
\begin{figure*}[t]
    \centering
    \includegraphics[width=0.8\linewidth]{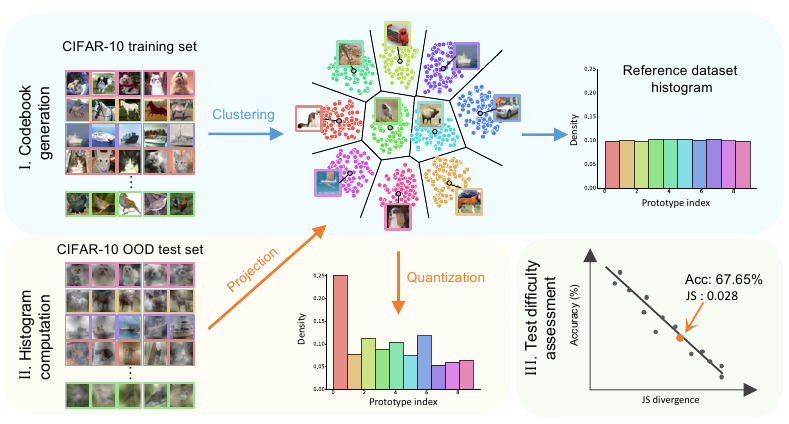}
    \vskip -0.1in
    \caption{Workflow of BoP representation computation using CIFAR-10 \cite{krizhevsky2009learning} and one CIFAR-10 out-of-distribution (OOD) test set as an example. \textbf{Top:} We group image features of the reference dataset CIFAR-10 into $10$ clusters, and the centers are called prototypes. The prototypes constitute the codebook of size $10$. \textbf{Bottom left:} To encode the OOD test set, we project it onto the codebook by quantizing each image feature to its corresponding prototype. Lastly, we compute the histogram, \emph{i.e.}, BoP representation, of CIFAR-10 OOD test set. \tu{\textbf{Bottom right:} We regard dataset-to-dataset similarity as the Jensen-Shannon divergence between BoP histograms of CIFAR-10 OOD test set and reference dataset. With such similarity, we can measure the test set difficulty for the model trained on reference dataset.}}
    \label{fig:flow}
    \vskip -0.2in
\end{figure*}

\subsection{Bag-of-Words Model Across Communities}
In \textbf{natural language processing} (NLP) and information retrieval, the Bag-of-Words (BoW) model \cite{mccallum1998comparison, joachims1998text, lewis1994sequential, joachims1997webwatcher} vectorizes textual data as a word histogram. Specifically, for each word in the dictionary, its occurrences in a document are counted, which fills in the corresponding entry of the BoW feature. This word frequency vector is thus used to represent a document. Numerous improvements of the BoW feature were made in NLP, such as n-grams \cite{lewis1994sequential, joachims1997webwatcher} and term frequency-inverse document frequency \cite{rajaraman2011mining}.

In the early 2000s, the BoW representation was introduced to the \textbf{computer vision} (CV) community to encode hundreds or thousands of local image descriptors \cite{lowe1999sift,bay2006surf} into a compact vector \cite{sivic2003bow}. As there is no semantic codebook like in NLP, a visual codebook is constructed by performing clustering (\eg, k-means) on a collection of local image features, where the resulting clustering centers are called ``visual words''. Local image descriptors are quantized to their nearest cluster center so that a visual word histogram can be computed. This BoW histogram also have undergone extensive improvements in later years, such as Fisher vector \cite{perronnin2007fisher, perronnin2010improving}, vector of locally aggregated descriptors (VLAD) \cite{jegou2010aggregating}, and the use of principal component analysis and whitening \cite{jegou2012negative}. 

\textbf{Contribution statement.} This paper contributes a baseline method in adopting the BoW idea study the two basic properties of a dataset. To this end, we propose to represent a dataset using its histogram over a series of prototypes. A comparison between the usage of BoW model in NLP, CV and our dataset-level research is shown in Table \ref{tab:analogy}. Specifically, the BoP representation relies on clustering for codebook formation, has a relatively small codebook (depending on the richness of dataset semantics), and has semantically sensible codewords.

\subsection{Bag-of-Prototypes Dataset Representation}

Given a dataset $\mathcal{D} = \{\mathbf{x}_i\}^N_{i = 1}$ where $N$ is the number of images and a feature extractor $\mathbf{F}(\cdot)$ that maps an input image into a $d$-dimensional feature $f \in \mathbb{R}^d$, we extract a set of image features $\mathcal{F} \coloneqq \{\mathbf{F}(\mathbf{x}_i)\}^N_{i=1}$. While it is possible to directly use the dataset images (or features) as model input under small $N$, it becomes prohibitively expensive when $N$ is large.  %
We therefore focus on extracting useful semantic features of $\mathcal{F}$ by encoding its image features into a compact representation. Below we detail the necessary steps for computing the proposed BoP representation (refer \cref{fig:flow}).

\textbf{Step I: Codebook generation.} 
Given a reference dataset $\mathcal{D}_r = \{\mathbf{x}^r_i\}^{N_{r}}_{i = 1}$, we extract all of its image features $\mathcal{F}_r \coloneqq \{\mathbf{F}(\mathbf{x}^r_i)\}^{N_{r}}_{i=1}$ using a pretrained network, from which a codebook is constructed. Specifically, we adopt standard k-means clustering \cite{macqueen1967some} to partition the feature space $\mathbb{R}^d$ into $K$ clusters. Each of the $K$ cluster centers is called a ``prototype'', because oftentimes each center mainly represents a certain semantic content. See \cref{fig:flow} right for exemplar image of each prototype. The prototypes, or centers, constitute the codebook, denoted %
as $\mathcal{C} = \{\mathbf{c}_i\}_{i=1}^K$, where $\mathbf{c}_i$ is the $i$-th prototype. Note that, the order of the prototypes is fixed in the codebook. %

\textbf{Step II: Histogram computation.}
{For a dataset to be encoded $\mathcal{D}_e = \{\mathbf{x}^e_i\}^{N_e}_{i = 1}$ where $N_{e}$ is the number of images, we project it onto codebook $\mathcal{C}$ of size $K$ to compute its BoP representation. Specifically, after extracting image features $\mathcal{F}_e \coloneqq \{\mathbf{F}(\mathbf{x}^e_i)\}^{N_{e}}_{i=1}$ from $\mathcal{D}_e$, for each image feature, we compute its distance with all the $K$ prototypes in the codebook, yielding $K$ distances $d_1,...,d_k$, where $d_i$ is the distance between an image feature and the $i$-th prototype. An image feature is quantized to prototype $c_i$ if $d_i$ is the lowest among $d_1,...,d_k$. Following the quantization, we generate a $K$-dimensional one-hot encoding where the $i$-th entry is 1 and all the others are 0. Having computed the one-hot vectors for all the image features, we sum them which is then normalized by $N_e$, the number of images in $D_e$. This gives the histogram representation $\mathbf{h}_{e}$, or \textbf{BoP representation}, for $D_e$ where the $i$-th entry indicates the density of features in $D_e$ belonging to prototype $\mathbf{c}_i$.}

\setlength{\tabcolsep}{2.0pt}
\begin{table}
\begin{center}
\small
\begin{tabular}{c|c|c|c}
	\Xhline{1.2pt} 	
    
                     & BoW in NLP                                   & BoW in CV                                          & BoP                                             \\ \hline
    \begin{tabular}[c]{@{}c@{}} Encoded \\ objects \end{tabular}          & Documents                                    & Images                                             & \begin{tabular}[c]{@{}c@{}} Datasets \\ (a set of images) \end{tabular}                                        \\ \hline
    \begin{tabular}[c]{@{}c@{}} Codewords in \\ codebook \end{tabular} & Words                                           & \begin{tabular}[c]{@{}c@{}} Cluster centers of \\ local descriptors\end{tabular}                       & \begin{tabular}[c]{@{}c@{}} Prototypes of \\ image features \end{tabular} \\ \hline
    Clustering? & No & Yes & Yes \\ \hline
    \begin{tabular}[c]{@{}c@{}} Codewords \\ semantics \end{tabular}  & Clear & Little & Sensible \\ \hline
    \begin{tabular}[c]{@{}c@{}} Codebook \\ size \end{tabular} & $>10^3$ & $10^3 - 10^6$ & \begin{tabular}[c]{@{}c@{}} $\sim10^2$ (dataset \\dependent) \end{tabular}\\ 	\Xhline{1.2pt} 	
\end{tabular}
\end{center}
\vskip -0.1in
\caption{Comparing BoP with BoW model in natural language processing (NLP) and computer vision (CV). The objective of BoW in NLP and CV is encoding texts and images respectively, while BoP is proposed to represent datasets.}
\vskip -0.1in
\label{tab:analogy}
\end{table}

\subsection{Measuring Dataset-to-Dataset Similarity}
\label{subsec:measure}
Similar to image / document retrieval where BoW vectors of instances are used for similarity comparison \cite{sivic2003bow, philbin2007object, csurka2004visual, nister2006scalable, fei2005baye}, this work uses the BoP representation to calculate dataset-to-dataset similarity. Specifically, given BoP representations $\mathbf{h}_x$ and $\mathbf{h}_y$ of two datasets $\mathcal{D}_x$ and $\mathcal{D}_y$, we simply define their similarity $S_{x,y}$ using Jensen-Shannon divergence (JS divergence), which is designed for histogram-based similarity measurement \cite{dagan-etal-1997-similarity, manning1999foundations}.

\textbf{Task-oriented similarity measure.} 
We can build a \textit{universal} codebook on a large-scale dataset following BoW model~\cite{csurka2004visual,zheng2017sift}.
By doing so, the resulting BoP representations are generic. 
We can also build a \textit{task-oriented} codebook on a reference dataset from a specific task to consider more task-oriented information.
The latter is more suitable for the two dataset-level tasks considered in this work.
For the task of training set suitability assessment, we use the target dataset as the reference for codebook generation to fully consider its the semantic information.
As a result, the JS divergence between BoP representations of the training set and the target dataset can well capture how a training set is similar to the target set.
Similarly, for the task of test set difficulty assessment, we build codebook on the training set. This practice can effectively measure how an unlabeled test is similar to a given training set.

\subsection{Discussion}
\textbf{Working mechanism of BoP.} 
Codebook generation of BoP can be viewed as Centroidal Voronoi Tessellations~\cite{du1999centroidal}. Specifically, the prototypes (cluster centers) of codebook tessellate the feature space into Voronoi cells. Then, histogram computation approximates a probability distribution function in the same way as the \textit{nonparametric} histogram~\cite{freedman1981histogram,boots2009spatial,polianskii2022voronoi}. That is, the BoP representation reflects the distribution of a dataset in the feature space.

As shown in Fig. \ref{fig:flow}, the prototypes of reference dataset tessellate feature space into Voronoi cells. Based on this, we quantify the histogram of the reference dataset to represent its distribution.
Given a new dataset, we conduct the same histogram calculation procedure and correspondingly capture its dataset distribution with the histogram.
Then, we measure discrepancy of the two datasets by calculating JS divergence between their histograms.
Compared with common measures of dataset distance (\eg, FD \cite{frechet1957distance}, KID \cite{binkowski2018demystifying} and MMD \cite{gretton2006kernel}) that only reflect global structure~(\eg, first few moments) of dataset distributions, BoP, collaborated with JS divergence, considers more local structures.

\textbf{Training set suitability vs. transferability estimation.} Two tasks relate but differ significantly: 1) Given an unlabeled target dataset and a pool of training datasets, the former aims to select the most suitable training set for the target. The latter assumes a labeled target dataset and a pool of models pretrained on a source dataset, with the goal of selecting the most suitable source model for the target without fine-tuning them all~\cite{pandy2022transferability, agostinelli2022stable, agostinelli2022transferability}; 
2) Datasets in training set suitability are used for the same classification problem. In contrast, in transferability estimation, the problem in the target dataset (\eg, CIFAR-10 classification) is different from that of the source dataset (\eg ImageNet classification).

\begin{figure*}[t]
    \centering
    \includegraphics[width=\linewidth]{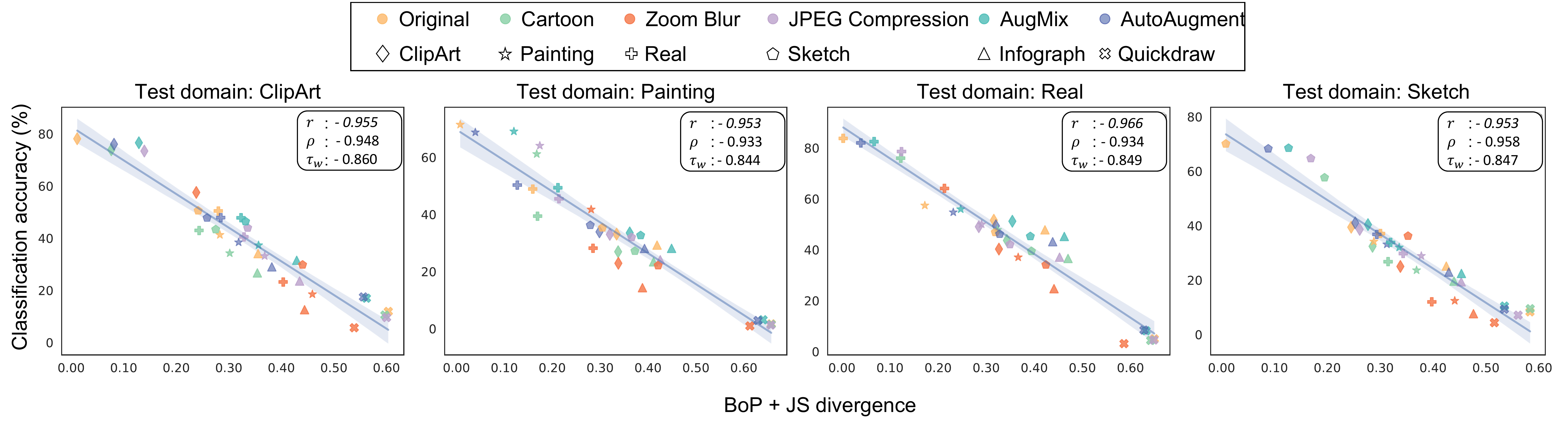}
    \vskip -0.1in
    \caption{\textbf{Correlation study for training suitability of datasets.}
    We report the correlation strength between BoP + JS and model classification accuracy on four test domains of DomainNet: ClipArt, Painting, Real and Sketch. \camera{The model architecture is ResNet-101.
    Each dot denotes a model trained on a training set of DomainNet. We mark training domains (\eg, ClipArt) by different shapes and transformation operations (\eg, AugMix) by different colors.}
    The straight lines are fit with robust linear regression \cite{huber2011robust}.
    We consistently observe high correlation results: Pearson's correlation ($\left|r\right| > 0.95$) , Spearman's correlation ($\left|\rho\right| > 0.93$) and weighted Kendall's correlation ($\left|\tau_w\right| > 0.84$). 
    This suggests that BoP + JS is predictive of the suitability of a training set.
    }
    \label{fig:domain_codebook}
    \vskip -0.2in
\end{figure*}

\textbf{Analysis of the number of prototypes in a codebook.}\label{dis:number}
The codebook size is a critical factor influencing the usefulness of the BoP. 
A small codebook means a coarser partition of feature space, where similar features will likely be in the same cluster, but dissimilar features may also be in the same cluster. Moreover, a large codebook provides a finer description of the space, where dissimilar features are quantized to different prototypes and more semantics are explored. According to our experiment results in \cref{fig:domain_codebook} and \cref{fig:codebook_size}, we find, reassuringly, BoP is robust against the codebook size: prototype number can deviate within a wide range around the true classes number (\eg, $345$ for DomainNet \cite{peng2019moment}) without significantly affecting performance.

\textbf{Application scope and future directions.} \label{sec:app_scope}
BoP is proposed to study the two dataset-level tasks, and the datasets considered in each task share the same label space.
We may encounter some situations where we need to compare datasets with different label spaces (\emph{e.g.}, pre-training datasets selection \cite{achille2019task2vec}).
In this case, one potential way is to build a universal codebook on a large-scale and representative dataset similar to BoW models~\cite{csurka2004visual,zheng2017sift}.
By doing so, the resulting BoP representations can encode diverse and sufficient semantics for comparing datasets across various label spaces.
We view our BoP as a starting point to encode datasets. It would be interesting to study other dataset vectorization schemes and dataset-level tasks.

\section{Comparing Training Suitability of Datasets}
This task studies dataset valuation where multiple training sets are provided by different data contributors. 
{The goal is to select the most suitable training set (ideally without training) whose trained model performs the best on a target test set.}
In this section, we first validate that BoP, collaborated with JS divergence (BoP + JS), is predictive of dataset suitability for the target test set. Then, we show that BoP is robust when using a wide range of codebook sizes and different networks.

\subsection{Experimental Settings} \label{sec:domain}
\textbf{Correlation study under DomainNet setup.} We use domain generalization benchmark DomainNet \cite{peng2019moment}, which consists of 6 domains: Painting, Real, Infograph, Quickdraw, Sketch and ClipArt, where the tasks are $345$-way object classification. Each domain has its training and test splits.
We conduct the correlation study in an \textit{leave-one-out} manner, leading to \textit{6 groups} of correlation studies, with each group using the test split of one domain as the target test set.
Additionally, we apply image transformations to the training split of six original domains. Specifically, we employ `\textit{Cartoon}' \cite{imgaug}, `\textit{Zoom Blur}' and `\textit{JPEG Compression}' \cite{hendrycks2019robustness} to convert domains' style to be one specific type. We also use `\textit{AugMix}' \cite{hendrycks2019augmix} and `\textit{AutoAugment}' \cite{cubuk2018autoaugment}, which transform images with various operations to generate domains with mixed styles. This process synthesizes $30$ new datasets, so we have $36$ training sets in total.

We follow the training scheme provided by TLlib \cite{tllib} to train ResNet-101 model \cite{He_2016_CVPR}, whose weights are pretrained on ImageNet \cite{deng2009imagenet}, yielding $36$ models.
Moreover, penultimate outputs of pretrained ResNet-101 is used as image feature.
On the test set, we generate a codebook of size $1000$. Then, for each training set, we compute its BoP histogram, BoP + JS from the test set, and the accuracy of its trained model on the test set. After this, we calculate correlation strength between BoP + JS and model accuracy to evaluate whether BoP is indicative of datasets training suitability. 

\textbf{Evaluation metric.} We use Pearson's correlation $r$ and Spearman's rank correlation $\rho$ to show linearity and monotonicity between BoP-based dataset distance and model accuracy, respectively. Both metrics range in $[-1, 1]$. If $\left|r\right|$ or $\left|\rho\right|$ is close to $1$, the linearity or monotonicity is strong, and vice versa.
In addition to these two metrics, we also use weighted variant of Kendall's correlation ($\tau_w$) \cite{vigna2015weighted}. It is shown to be useful when selecting the best ranked item is of interest \cite{shieh1998weighted}, while a major application of BoP + JS is to {select the training dataset leading to the best performance on a test set}. This metric has the same range where a number closer to $-1$ or $1$ indicates stronger negative or positive correlation, respectively, and $0$ means no correlation.

\subsection{Evaluation}

\setlength{\tabcolsep}{3pt}
\begin{table}

\begin{center}
\normalsize
\begin{tabular}{cc ccc ccc}
	\toprule
\multirow{2}{*}{\textbf{Method}}  & \multicolumn{3}{c}{\textbf{ResNet-34}} & \multicolumn{3}{c}{\textbf{ResNet-101}}\\
\cmidrule(lr){2-4}  \cmidrule(lr){5-7}
& $r$ & $\rho$ & $\tau_w$ & $r$ & $\rho$ & $\tau_w$ \\
		\midrule
		FD \cite{frechet1957distance} & -0.860 & -0.926 & -0.828 
               & -0.903  & -0.902 & -0.802  \\ 
		MMD \cite{gretton2006kernel} & -0.817 & -0.801 & -0.691
                & -0.821  & -0.817 & -0.704 \\ 
		KID \cite{binkowski2018demystifying} & -0.773  & -0.904 & -0.804
                & -0.876  & -0.896 & -0.800  \\ 
		\cmidrule(lr){1-7}
		BoP + JS  & \textbf{-0.960} & \textbf{-0.927} & \textbf{-0.840}
                      & \textbf{-0.961} & \textbf{-0.929} & \textbf{-0.840}\\
            \bottomrule
\end{tabular}
\end{center}
\vskip -0.15in
\caption{Compare averaged Pearson's correlation ($r$), Spearman's correlation ($\rho$) and weighted Kendall's correlation ($\tau_w$) of Fr\'echet distance (FD), maximum mean discrepancy (MMD) , kernel inception distance (KID) and BoP + JS (codebook size $1000$) on six test sets in DomainNet.
We report two groups of results using ResNet-34 (Left) and ResNet-101 (Right).
We show BoP + JS is more effective in assessing training set suitability than others.
}
\vskip -0.1in
\label{tab:domain_arch}
\end{table}

\textbf{Strong correlation: A training set is more suitable for a given test set if it has small BoP + JS.} 
\cref{fig:domain_codebook} shows correlation study on ClipArt, Painting, Real and Sketch. We notice that there are strong Pearson's correlations ($\left|r\right| > 0.95$), Spearman's rank correlations ($\left|\rho\right| > 0.93$) and relatively high weighted Kendall's correlations ($\left|\tau_w\right| > 0.84$) on four test sets. This suggests that BoP + JS is stable and useful across test sets. Table \ref{tab:domain_arch} compares average correlation strength of BoP + JS with Fr\'echet distance (FD) \cite{frechet1957distance}, maximum mean discrepancy (MMD) \cite{gretton2006kernel} and kernel inception distance (KID) \cite{binkowski2018demystifying}. \camera{They use that same image features as BoP. According to their formulae, mean and covariance of these features are used for distance computation.} We see that BoP + JS has the highest average correlation scores on six test sets ($\left|r\right| = 0.961$, $\left|\rho\right| = 0.929$ and $\left|\tau_{w}\right| = 0.840$). On average, BoP + JS is superior in depicting training sets suitability for a test set without any training procedure.

\begin{figure}
    \centering
    \includegraphics[width=\linewidth]{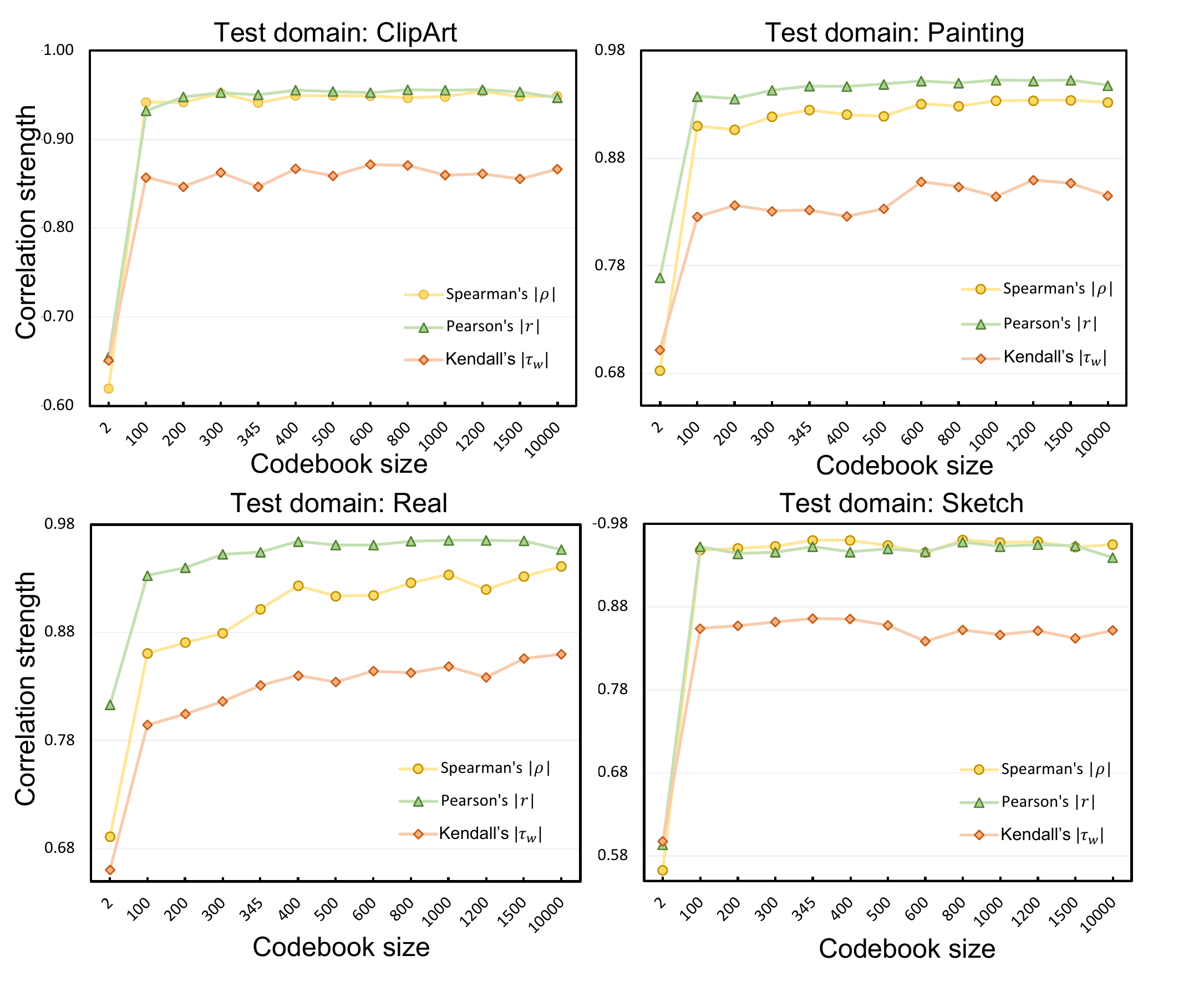}
    \vskip -0.1in
    \caption{\tu{\textbf{Impact of codebook size on correlation strength for ResNet-101 on four test domains: ClipArt, Painting, Real and Sketch.}
    For example, on Real domain, correlation scores $\left|\rho\right|$, $\left|r\right|$ and $\left|\tau_w\right|$ are relatively low under a small size and remain stably high when the size is greater than $400$.}}
    \label{fig:domain_codebook_size}
    \vskip -0.2in
\end{figure}

\textbf{Impact of codebook size} is shown in the \cref{fig:domain_codebook_size}. We construct codebooks with different size within approximately one order of magnitude around $345$. We find that the three correlation scores increase and then become stable when codebook size becomes larger. This indicates that the performance BoP + JS is overall consistent. 

\textbf{Correlation study with a different model architecture.} We additionally validate the robustness of BoP for ResNet-34 with codebook size $1000$. As shown in Table~\ref{tab:domain_arch}, we compare the average correlation scores of BoP + JS, FD, MMD and KID. We see that BoP + JS has consistent performance on two models and remains preferable to characterize training suitability.

\section{Assessing Test Set Difficulty without Labels}
In the task of test set difficulty assessment, we are provided with a labeled training set and a set of unlabeled datasets for testing.
Given a classifier trained on the training set, the goal is to estimate the model accuracy on these test sets without any data annotations.
In this section, we first show dataset distance measured by BoP + JS exhibits strong negative correlation with classifier accuracy. We then demonstrate an accuracy predictor based on the BoP representation gives accurate performance estimates compared to state-of-the-art methods.

\begin{figure*}
    \centering
    \includegraphics[width=\linewidth]{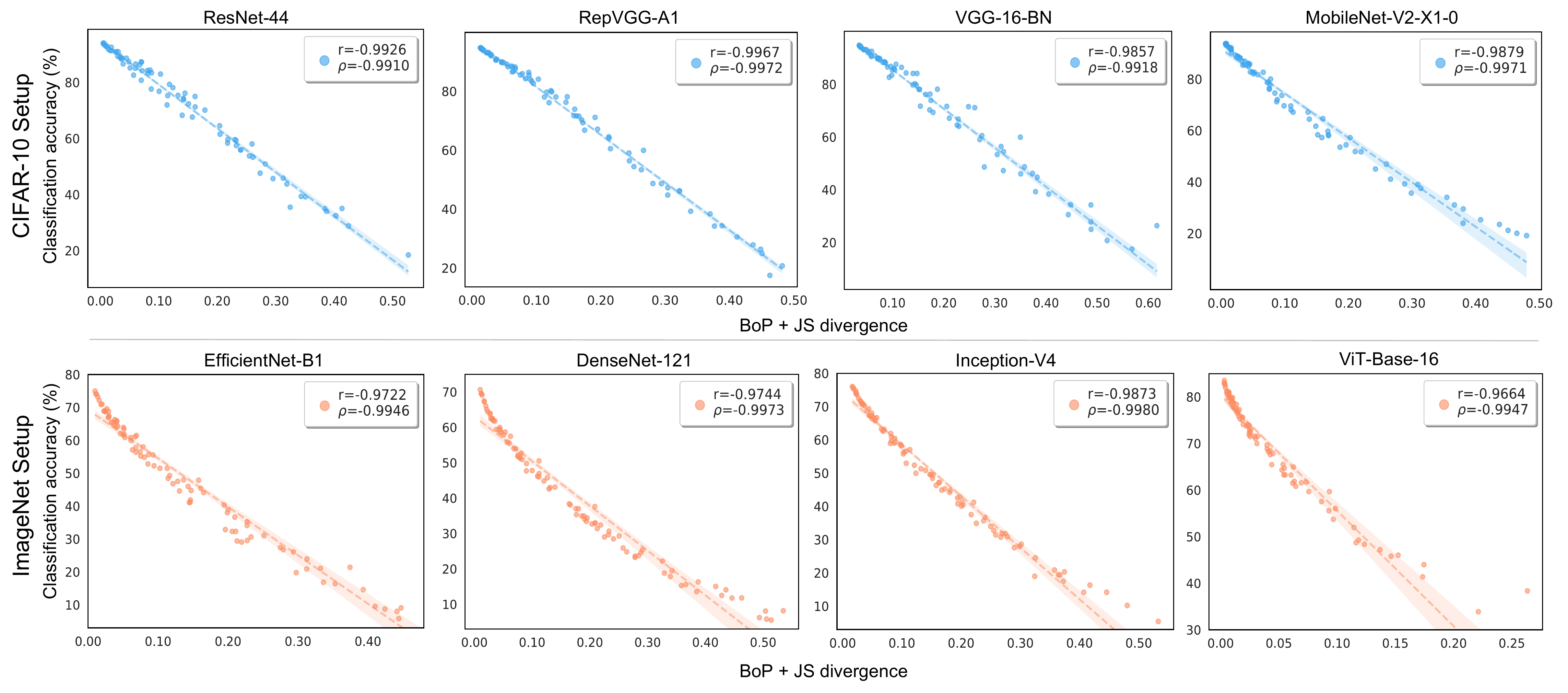}
    \vspace{-0.6cm}
    \caption{\textbf{Correlation between train-test distance measured by BoP + JS and model accuracy.}
    \textbf{Top:} Correlation study under CIFAR-10 setup using ResNet-44, RepVGG-A1, VGG-16-BN and MobileNet-V2. Each data point denotes a dataset from CIFAR-10-C. 
    \textbf{Bottom:} Correlation study under ImageNet setup using EfficientNet-B1, DenseNet-121, Inception-V4 and ViT-Base-16. ImageNet-C datasets are used as test sets. The straight lines are fit with robust linear regression \cite{huber2011robust}.
    Under both setups, we observe the strong Spearman's rank correlation ($\left|\rho\right| > 0.98$) between BoP + JS and model accuracy.
    }
    \vspace{-0.6cm}
    \label{fig:correlation}
\end{figure*}

\subsection{Experimental Settings} \label{sec:settings-difficulty}
\textbf{Correlation study under CIFAR-10 setup.} We conduct a correlation study by comparing BoP + JS with classifier accuracy. Following the same setup in \cite{deng2021pami}, we use a series of datasets sharing the same label space (but usually with distribution shift) with CIFAR-10 \cite{krizhevsky2009learning}. %
Specifically, we train {{ResNet-44}} classifier \cite{He_2016_CVPR} on the training set of CIFAR-10, which consists of $50,000$ images from $10$ classes. 
Here, we use the {{CIFAR-10-C benchmark}} \cite{hendrycks2019benchmarking} for correlation study, which contains different types of corruptions with $5$ levels of severity including {per-pixel noise, blurring, synthetic weather effects, and digital transforms}.  %
Then, for each dataset, we compute its BoP vector, its BoP + JS from CIFAR-10 training set and the classifier accuracy. In addition to ResNet-44, we also study the RepVGG-A1 \cite{ding2021repvgg}, VGG-16-BN \cite{simonyan2014very} and MobileNet-V2 \cite{sandler2018mobilenetv2} . 

\textbf{Predicting classification accuracy under CIFAR-10 setup.} 
We train a regressor that takes as input the BoP representation and outputs classification accuracy. 
The regressor is a neural network with $3$ fully connected layers and trained on CIFAR-10-C (regression training set). %
We evaluate accuracy prediction on CIFAR-10.1 \cite{recht2018cifar}, CIFAR-10.2 \cite{recht2018cifar} and CIFAR-10.2-$\Bar{C}$ \cite{mintun2021interaction}.
The former two are real-world datasets with natural shift, while the latter one is manually corrupted by the same synthetic shift as \cite{mintun2021interaction}.  %
Specifically, we add $10$ types of unseen and unique corruptions such as {warps}, {blurs}, {color distortions} and {noise additions}, with $5$  severity levels to CIFAR-10.2. 
Note that, these corruptions have no overlap with those in CIFAR-10-C \cite{mintun2021interaction}.

For the above, we extract image features (output of penultimate layer of ResNet-44) from CIFAR-10 training set. %
We construct a codebook by dividing the features into $80$ clusters with k-means.

\textbf{Correlation study under ImageNet setup.} %
We use DenseNet-121 \cite{huang2017densely} classifier trained on ImageNet training set. We employ a series of datasets from the ImageNet-C benchmark \cite{hendrycks2019robustness} where the classifier is tested. ImageNet-C uses the same types of corruptions as CIFAR-10-C.
We construct a codebook of size $1000$ on the ImageNet training set from which images features are extracted by the penultimate layer of DenseNet-121.
We project each dataset in ImageNet-C onto the codebook and obtain their BoP representations. When exhibiting linear correlations, we calculate BoP + JS between each ImageNet-C dataset and the training set, and compute classification accuracy.  We also use EfficientNet-B1 \cite{tan2019efficientnet}, Inception-V4 \cite{szegedy2017inception} and ViT-Base-16 \cite{dosovitskiy2020vit} to repeat above procedure for correlation study.

\textbf{Evaluation metric.} Same as Section \ref{sec:domain}, we use Pearson's correlation $r$ and Spearman's rank correlation $\rho$ to show linearity and monotonicity between BoP based dataset distance and model accuracy, respectively. 
To evaluate the effectiveness of accuracy estimation, we use root mean squared error (RMSE) by calculating the difference between estimated accuracy and ground truth before taking the mean across all the test sets. A larger RMSE means a less accurate prediction, and vice versa.

\setlength{\tabcolsep}{4.5pt}
\begin{table*}[t]
    \caption{
    Method comparison in predicting classifier accuracy under CIFAR-10 setup. We compare four methods: predicted score-based method with hard threshold $\tau$, neural network regression based on feature statistics ($\mu + \sigma + $FD) \cite{deng2021labels}, average thresholded confidence with maximum confidence score function (ATC-MC) \cite{garg2022leveraging} and difference of confidences (DoC) \cite{guillory2021predicting}.
    We use CIFAR-10.1 and CIFAR-10.2 (both real-world) and CIFAR-10.2-$\Bar{C}$ (manually corrupted)  as unseen test sets for accuracy prediction.
    We use RMSE (\%) to indicate precision of estimates. In each column, we compare our method with the best of the competing ones. 
    We report results by average of five runs.
    } 
\small
    \vspace{-0.4cm}
    \begin{center}
    \begin{tabular}{l|cc|c c c c c|c}
    \toprule
    \multirow{2}{*}{Method} & \multirow{2}{*}{CIFAR-10.1} & \multirow{2}{*}{CIFAR-10.2} & \multicolumn{6}{c}{CIFAR-10.2-$\Bar{C}$ (50)}\\
    \cline{4-9}
    & & & Severity $1$ & Severity $2$ & Severity $3$ & Severity $4$ & Severity $5$ & Overall\\
\midrule
    Prediction score ($\tau = 0.8$) & $4.899$ & $4.800$ & $10.127$ & $12.869$ & $16.809$ & $21.427$ & $24.371$ & $17.910$\\
    Prediction score ($\tau = 0.9$)  & $0.297$ & $0.550$ & $3.638$ & $5.078$ & $8.048$ & $11.804$ & $14.108$ & $9.404$ \\
     \midrule 
    ATC-MC \cite{garg2022leveraging} & $2.650$ & $2.672$ & $3.080$ & $ 4.306$ & $7.108$ & $11.015$ & $13.040$ & $8.601$ \\
    \midrule
    DoC \cite{guillory2021predicting} & $0.490$ & $0.263$ & $\mathbf{2.247}$ & $2.916$ & $5.117$ & $9.012$ & $6.637$ & $5.745$\\
   \midrule
    $\mu + \sigma + FD$ \cite{deng2021pami} & $0.455$ & $0.561$ & $5.875$ & $5.823$ & $4.724$ & $4.908$ & $6.486$ & $5.602$\\
\midrule
    BoP ($K=80$) & $0.218$ & $\mathbf{0.122}$ & $2.458$ & $2.818$ & $3.730$ & $5.836$ & $6.451$ & $4.551$ \\
    {BoP (${K=100}$)} & $\mathbf{0.186}$ & $0.124$ & $2.849$ & $\mathbf{2.808}$ & $\mathbf{3.548}$ & $\mathbf{4.025}$ & $\mathbf{4.777}$ & $\mathbf{3.677}$\\
\bottomrule
    \end{tabular}
    \end{center}
    \label{tab:autoeval}
\end{table*}

\begin{figure*}
    \centering
    \vskip -0.2in
    \includegraphics[width=\linewidth]{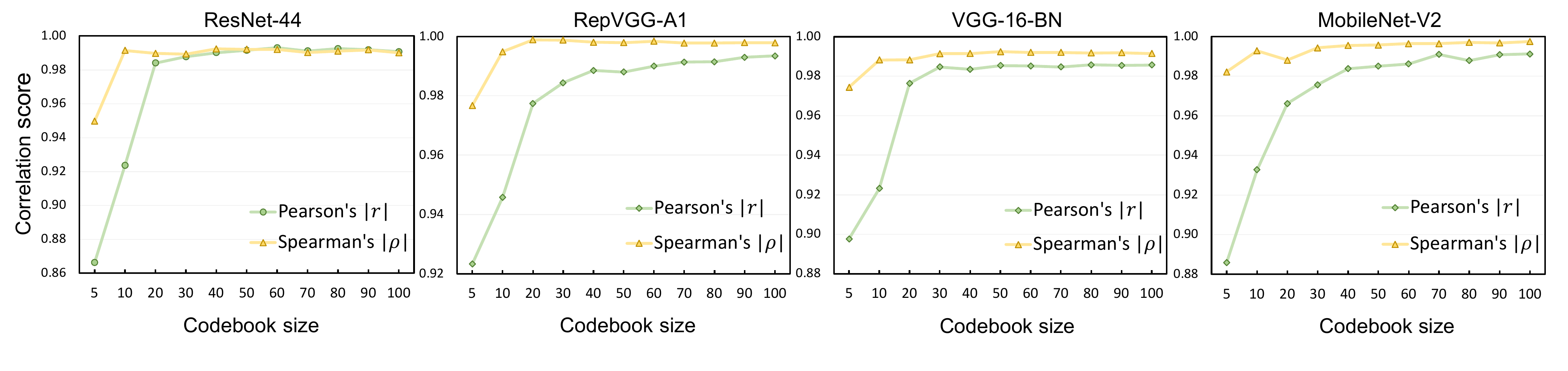}
    \vskip -0.15in
    \caption{\tu{\textbf{Impact of codebook size on correlation strength on CIFAR-10-C.}
    Correlation scores $\left|\rho\right|$ and $\left|r\right|$ are relatively low under a small size and become stably high when the size is greater than $20$ for all four model architectures.}}
    \label{fig:codebook_size}
    \vskip -0.2in
\end{figure*}

{\textbf{Compared methods.}} We compare our system with four existing ones. 1) \textit{Prediction score:} %
it estimates model accuracy using the maximum of Softmax output (\emph{i.e.}, confidence score). An image with a confidence score greater than a predefined threshold $\tau \in [0, 1]$ is considered correctly predicted. We select two thresholds ($\tau =0.8$ and $0.9$). 2) \textit{Difference of confidence (DoC)} \cite{guillory2021predicting} trains a linear regressor mapping average confidence to classifier accuracy; 3) \textit{Average thresholded confidence with maximum confidence score function (ATC-MC)} \cite{saurabh2022atc} calculates a threshold on CIFAR-10 validation set and regards an image with a confidence score higher than the threshold as correctly classified; 4) \textit{Network regression ($\mu + \sigma + $FD)} \cite{deng2021pami} trains a neural network that takes as input the feature mean, covariance and Fr\'echet distance between a set of interest and training set and outputs model accuracy. All methods, if applicable, are compared under the same conditions as our system, \eg, classification training set and regression training set.

\subsection{Evaluation}

{\textbf{Strong correlation: A test set is difficult (low accuracy) if it is dissimilar to the training set using  BoP + JS.} \cref{fig:correlation} presents the correlation study of two setups and various classifiers. %
We observe a very high Spearman's rank correlation ($\left|\rho\right| > 0.99$) in all the scenarios. It indicates that classification accuracy is highly correlated with JS divergence between BoPs of training and test sets. That is, test accuracy drops proportionally to the distance between the given training set and a test set.
The results demonstrate BoP + JS between training and test sets is an effective indicator of classification accuracy. More studies are presented in the supplementary materials. 
}

\textbf{Effectiveness of the BoP representation in predicting classification accuracy on variou unseen test sets.} 
After performing correlation study, we train a neural network regressor on CIFAR-10-C and test it on a series of test sets. Results are summarized in Table \ref{tab:autoeval}. 
We have the following observations. \textbf{First and foremost}, BoP representation achieves the best accuracy prediction performance, evidenced by the lowest RMSE across all the four test scenarios. For example, on the test sets of CIFAR-10.2-$\Bar{C}$, the RMSE of our method is $4.551$, which is $1.051$ lower than the second best method \cite{deng2021pami}. This clearly validates the effectiveness of the BoP representation. \textbf{Second}, we observe that the ``Prediction score'' method is unstable. While it has good results under $\tau=0.9$ on CIFAR-10.1 and CIFAR-10.2 datasets, it is generally inferior to the competing methods in other test scenarios. Our observation is similar to \cite{deng2021pami}, suggesting that a more robust threshold selection method is needed for this method. \textbf{Third,} although BoP has slightly higher RMSE than DoC on Severity $1$ of CIFAR-10.2-$\Bar{C}$ ($2.458$ \emph{v.s.}, $2.247$), we stress that BoP is overall more stable and effective on real world datasets and other severity levels of synthetic datasets.

\textbf{Impact of codebook size} is summarized in \cref{fig:codebook_size} under CIFAR-10 setup.
We conduct the study using different sizes on four classifiers. We observe correlation scores first increase and then become stable when codebook size is larger than $20$. These results are considered validation and help us use competitive and stable codebook sizes in Table \ref{tab:autoeval}.   %

\vskip -0.3in
\section{Conclusion}
This work introduces a \textit{bag-of-prototypes} (BoP) dataset representation to vectorize visual datasets. 
It first computes a codebook composed of clustering prototypes and then a prototype histogram for a dataset.
The BoP vector considers the underlying local semantic distribution of a dataset and is thus more discriminative than global dataset statistics. 
Specifically, when used in conjunction with JS divergence, the proposed descriptor better captures the underlying relationship across datasets. This advantage is validated by its promising results in two dataset-level tasks: assessing training set suitability and test set difficulty. 
This work establishes the baseline usage of the BoP scheme, and more investigations and applications will be made in future work. 

\section*{Acknowledgements} 
We thank all anonymous reviewers for their constructive comments in improving this paper. This work was supported by the ARC Discovery Project (DP210102801).

\newpage

{\small
\bibliographystyle{ieee_fullname}
\bibliography{egbib}
}

\cleardoublepage

\appendix
\section{Supplementary}
In this supplementary material, we first introduce the experimental details including the training schemes, datasets and computation resources.
Then, we show the full results of the correlation study on the task: training set suitability in \cref{fig:all} and impact of codebook size in \cref{fig:codebook}. Last, we report the accuracy estimation results on each severity level of CIFAR-10.1-$\Bar{C}$ in Table \ref{tab:10.1} and results on datasets with natural distribution shifts in \cref{fig:real}. After that, we present a correlation study of average thresholded confidence, average confidence, and difference of confidence on ImageNet and CIFAR-10 setups.
% \end{abstract}

%%%%%%%%% BODY TEXT
\subsection{Experimental Setup}
\subsubsection{Datasets} 
We carefully check the licenses of all datasets used in the experiment and list the open sources to them.

\noindent \textbf{DomainNet} \cite{peng2019moment} (\textcolor{blue}{http://ai.bu.edu/M3SDA/});\\
\textbf{ImageNet} \cite{deng2009imagenet} (\textcolor{blue}{https://www.image-net.org}); \\
\textbf{ImageNet-C} \cite{hendrycks2019robustness} (\textcolor{blue}{https://github.com/hendrycks/robustness});\\
% \textbf{ImageNet-C-Bar} (\textcolor{blue}{https://github.com/facebookresearch/augmentation-corruption});\\
\textbf{CIFAR-10} \cite{krizhevsky2009learning} (\textcolor{blue}{https://www.cs.toronto.edu/~kriz/cifar.html});\\
\textbf{CIFAR-10-C} \cite{hendrycks2019robustness} (\textcolor{blue}{https://github.com/hendrycks/robustness}); \\
\textbf{CIFAR-10.1} \cite{recht2018cifar} (\textcolor{blue}{https://github.com/modestyachts/CIFAR-10.1});\\
\textbf{CIFAR-10.2} \cite{recht2018cifar} (\textcolor{blue}{https://github.com/modestyachts/CIFAR-10.1});\\
\textbf{CIFAR-10-$\Bar{C}$} \cite{mintun2021interaction} (\textcolor{blue}{https://github.com/facebookresearch/
augmentation-corruption}) We use the corruption method provided in this link to create CIFAR-10.1-$\Bar{C}$ and CIFAR-10.2-$\Bar{C}$;

\begin{figure*}[t]
    \centering
    \includegraphics[width=0.88\linewidth]{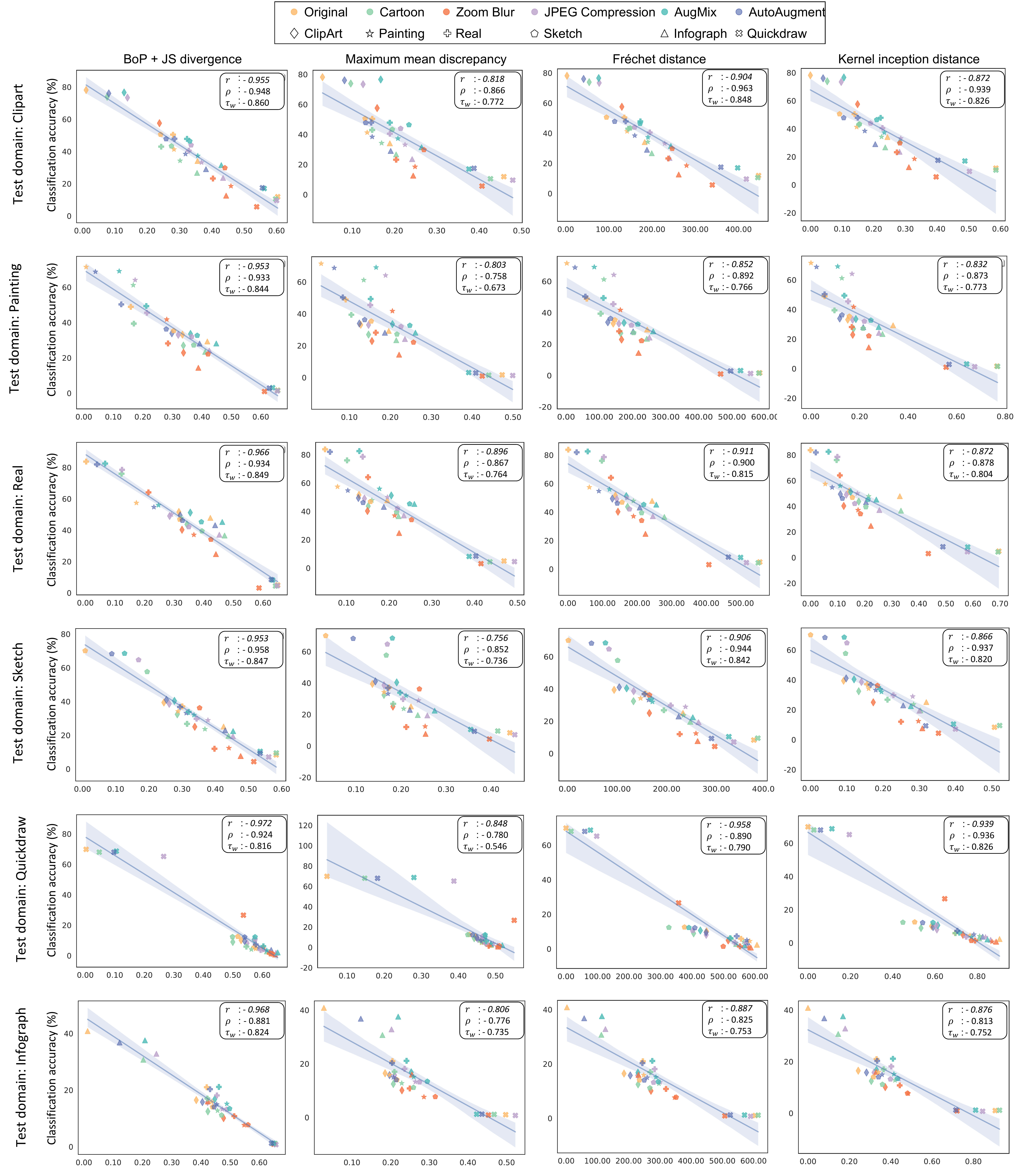}
    \caption{\textbf{The full results of comparing training suitability of datasets.} Each point denotes a model. The models trained on transformed training sets are marked with shapes and each shape denote one specific transformation operation (\eg, AutoAugment \cite{cubuk2018autoaugment}).
    The straight lines are fit with robust linear regression \cite{huber2011robust}.}
    \label{fig:all}
    % \vskip -0.12in
\end{figure*}

\subsubsection{Experiment: Datasets Training Suitability}
Follow the same training scheme as \cite{tllib}, we use ResNet-101 \cite{He_2016_CVPR} architecture pre-trained on ImageNet \cite{deng2009imagenet}. The training epoch is $20$, the batch size is 32 and the number of iterations per epoch is $2500$. The optimizer is SGD with learning rate $1 \times 10^{-2}$ and weight decay $5 \times 10^{-4}$.

\subsubsection{Experiment: Datasets Testing Difficulty}
\textbf{CIFAR-10 setup.} We use ResNet-44, RepVGG-A1, VGG-16-BN and MobileNet-V2 classifiers and their trained weights are publicly released by \textcolor{blue}{https://github.com/chenyaofo/pytorch-cifar-models}.

\textbf{ImageNet setup.} We use EfficientNet-B1, DenseNet-121, Inception-V4 and ViT-Base-16 classifiers for correlation study. The pretrained models are provided by PyTorch Image Models (timm) \cite{rw2019timm}.

\subsubsection{Computation Resources} 
PyTorch version is \textcolor{black}{1.10.2+cu102} and timm version is \textcolor{black}{1.5}. 
All experiments run on one 2080Ti and the CPU \textcolor{black}{AMD Ryzen Threadripper 2950X 16-Core Processor}.

\subsection{Results of Datasets Training Suitability}
\textbf{Full results of correlation study on six test domains.} We present a correlation study of BoP + JS divergence, Fr\'echet distance, maximum mean discrepancy and kernel inception distance. We use ResNet-101 as the feature extractor. The codebook size for BoP is $1000$. All methods use the same feature. Based on their formulae, we use mean and covariance feature to compute them. On all domains, we see that BoP has consistent and superior performance.

\begin{figure*}
    \centering
    \includegraphics[width=0.88\linewidth]{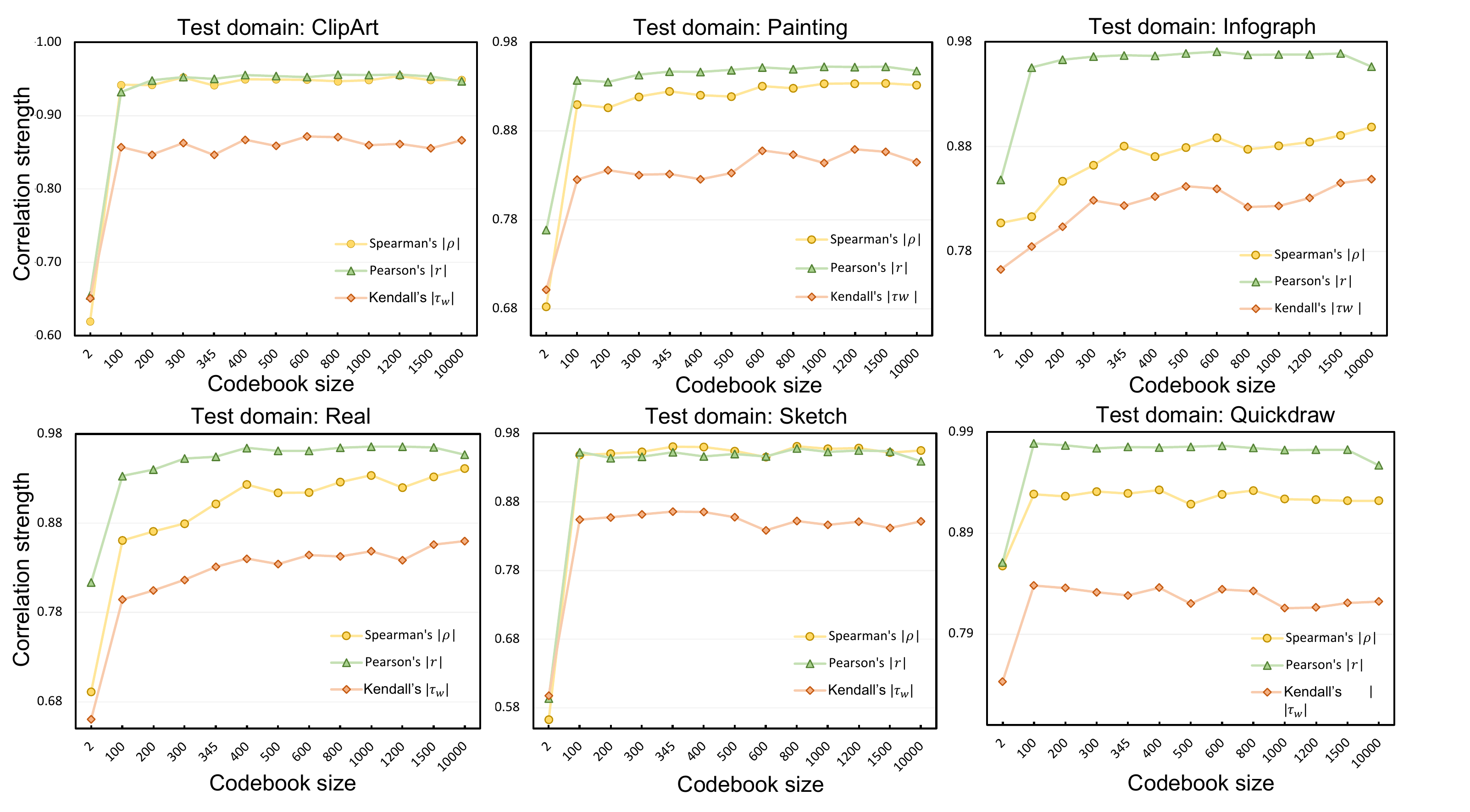}
    \caption{\textbf{The impact of codebook size on correlation strength on six test domains: ClipArt, Painting, Infograph, Real, Sketch and Quickdraw.} We observe on six domains that BoP+JS gives low correlation with a small codebook and maintains stably high when codebook size becomes larger.}
    \label{fig:codebook}
    % \vskip -0.12in
\end{figure*}

\textbf{Impact of codebook size on correlation strength for ResNet-101.} In \cref{fig:codebook}, we find that BoP + JS divergence gives a relatively low correlation and converges to a high correlation when codebook size becomes larger.

\textbf{Use Hellinger and Chi-squaured to measure the distance between BoP representations.} We test other distances than JS, such as Hellinger and Chi-squaured under DomainNet setup with ResNet-101 and codebook size $1000$. They yield similar results ($\left|\rho\right| = 0.928, 0.927$) as JS ($\left|\rho\right| = 0.929$). These results further validate the usefulness of BoP.

\textbf{Use ResNet-34 for model accuracy and ResNet-101 to extract features.} We use the extracted features to construct a codebook size $1000$ and JS divergence to measure distances. BoP + JS divergence gives $\left|\rho\right| = 0.927$, and FD, MMD and KID gives $\left|\rho\right| = 0.909, 0.825, 0.899$, respectively. For Pearson's correlation, BoP + JS is still the highest ($\left|\rho\right| = 0.959$) compared to FD, MMD and KID ($\left|\rho\right| = 0.898, 0.823, 0.864$).

\subsection{Results of Datasets Testing Difficulty}

\setlength{\tabcolsep}{2pt}
\begin{table}

\begin{center}
\normalsize
\begin{tabular}{cc ccc ccc}
	\toprule
\multirow{2}{*}{\textbf{Method}}  & \multicolumn{6}{c}{\textbf{CIFAR-10.1-$\Bar{C}$}} \\
\cmidrule(lr){2-7} 
& L$1$ & L$2$ & L$3$ & L$4$ & L$5$ & Overall \\
		\midrule
		$\tau = 0.8$ & 10.43 & 13.02 & 16.44 & 21.35 & 24.44 & 17.90\\ 
		$\tau = 0.9$ & 4.00 & 5.26 & 8.36 & 11.90 & 13.98 & 9.49\\
       \midrule  
		ATC-MC \cite{garg2022leveraging} & 3.47 & 4.63 & 7.64 & 11.15 & 12.85 & 8.73\\ 
            DoC \cite{guillory2021predicting} & 2.48 & 2.90 & 6.30 & 8.80 & 6.87 & 5.98\\
        \midrule  
            $\mu + \sigma + FD$ \cite{deng2021pami} & 6.46 & 6.12 & 5.51 & 4.58 & 5.52 & 5.67\\
        \midrule
          BoP ($K=80$) & 1.89 & 2.34 & 3.56 & 5.92 & 6.32 & 4.40\\
          BoP ($K=100$) & 2.14 & 2.83 & 3.98 & 3.68 & 5.52 & 3.81\\
            \bottomrule
\end{tabular}
\end{center}
% \vskip -0.15in
\caption{Method comparison in predicting classifier accuracy under CIFAR-10 setup. We report RMSE (\%) on each severity level of CIFAR-10.1-$\Bar{C}$.}
% \vskip -0.1in
\label{tab:10.1}
\end{table}

\textbf{Full results of correlation study on CIFAR-10.1-$\Bar{C}$} is shown \cref{fig:all}. We see that BoP + JS divergence consistently well correlates with classification accuracy on six test domains: ClipArt, Painting, Real, Sketch, Quickdraw, Infograph with $\left|r\right| > 0.95$, $\left| \rho \right| > 0.88$ and $\left| \tau_w \right| > 0.81$.

\begin{figure*}[t]
    \centering
    \includegraphics[width=0.88\linewidth]{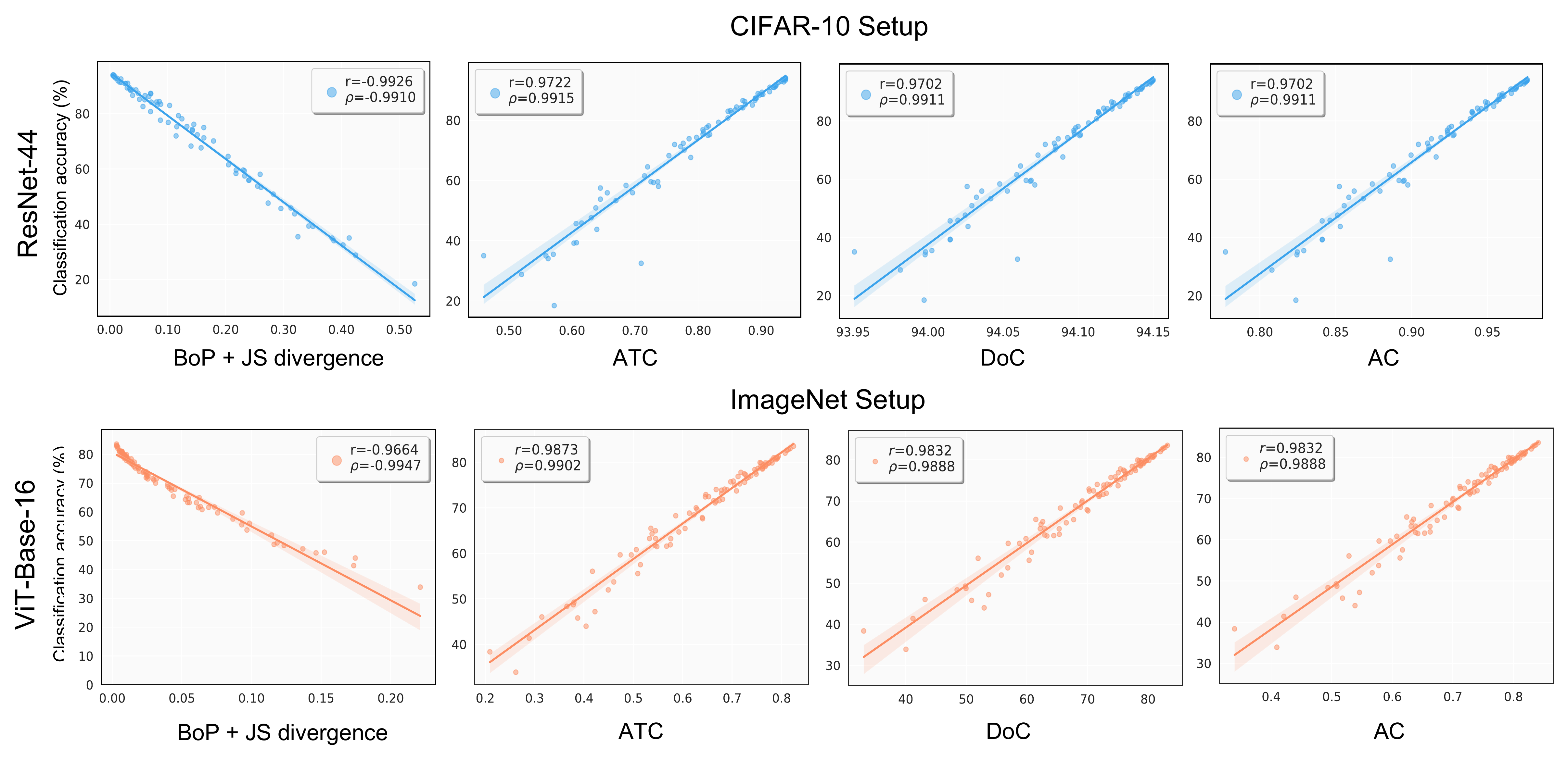}
    \caption{\textbf{The correlation study of BoP + JS, ATC, DoC and AC under CIFAR-10 and ImageNet setups.} \textbf{Top:} Correlation study under CIFAR-10 setup using ResNet-44. In each figure, a point denotes a dataset from CIFAR-10-C benchmark. \textbf{Bottom:} Correlation study under ImageNet setup using ViT-Base-16. In each figure, a point denotes a dataset from ImageNet-C benchmark.
    The straight lines are fit with robust linear regression \cite{huber2011robust}. We observe that BoP+JS gives a higher Spearman's correlation on both setups.}
    \label{fig:supp}
    % \vskip -0.12in
\end{figure*}

\textbf{The correlation study of BoP + JS, ATC, DoC and AC under CIFAR-10 and ImageNet setups for ResNet-44 and ViT-Base-16, respectively.} We include results of (1) Prediction score ($\tau = 0.8$, $\tau = 0.9$), (2) Difference of confidence (DoC)\cite{guillory2021predicting}, (3)Average thresholded confidence with maximum confidence (ATC-MC) \cite{garg2022leveraging}, Network regression ($\mu + \sigma + FD$) \cite{deng2021pami} and BoP with codebook size $80$ or $100$ on CIFAR-10.1-$\Bar{C}$ in Table \ref{tab:10.1}. We see that BoP is overall more predictive of model testing difficulty compared with other methods. We also present the correlation study of BoP + JS, ATC-MC, DoC and average confidence (AC) under CIFAR-10 setup and ImageNet setup in \cref{fig:supp}.

\begin{figure}
    \centering
    \vskip -0.1in
    \includegraphics[width=\linewidth]{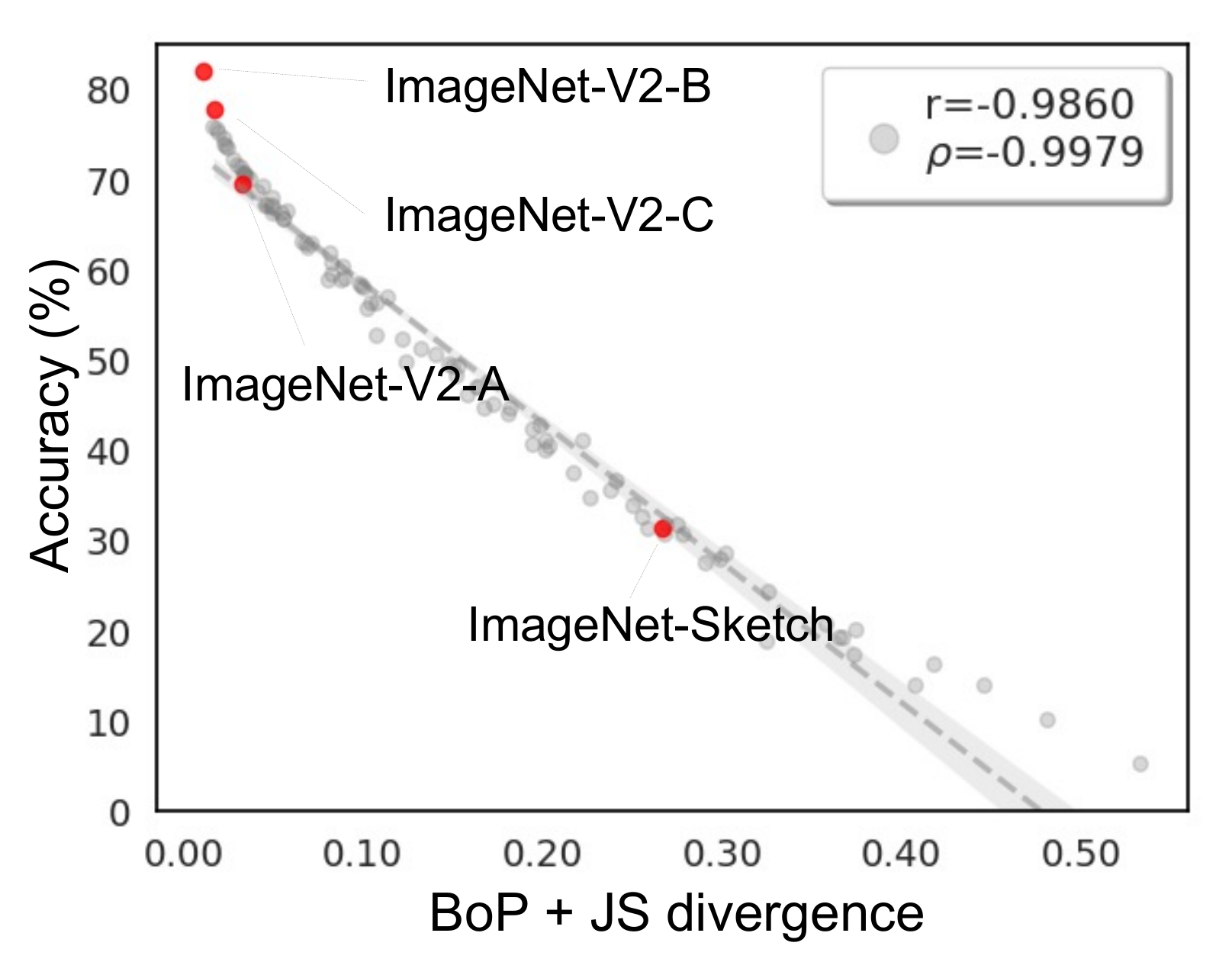}
    \vskip -0.1in
    \caption{\textbf{Results on four datasets with natural shifts for Inception-V4: ImageNet-V2-A/B/C and ImageNet-Sketch.}
    % of Spearman's rank correlation ($\left|\rho\right|$) and Pearson correlation ($\left|r\right|$). 
    % We test ResNet-44, RepVGG-A1 and VGG-16-BN.
    We find that four new datasets (red dots) still lie on the original trend of ImageNet-C (grey datasets).}
    % Study on the impact of codebook size on  }}
    \label{fig:real}
    \vskip -0.2in
\end{figure}

\textbf{Results on datasets with natural distribution shifts for Inception-V4.} In addition to datasets with synthetic shifts (\ie ImageNet-C), we explore the effectiveness of BoP on datasets with natural shifts. Under ImageNet setup, we include results of ImageNet-V2-A/B/C and ImageNet-Sketch for Inception-V4. From \cref{fig:real}, we observe that four datasets (red dots) still lie on the trend of ImageNet-C (grey dots). This means that BoP still effectively captures the distributional shift of four real-world datasets.

%%%%%%%%%%%%%%%%%%%%%%%%%%%%%%%%%%%%%%%%%%%%%%%%%%%%%%%%%%%%

\textbf{DomainNet setup for test set difficulty.} We also use $36$ domains from DomainNet setup for datasets testing difficulty. We use ResNet-101 trained on `Real' domain to extract features and construct a codebook of size $1000$. We conduct a correlation study between BoPs of the rest $35$ domains and model accuracy. We compare BoP + JS with ATC, DoC and AC. We find that BoP + JS shows the strongest correlation ($\left|r\right| = 0.959, \left|\rho\right| = 0.972$) while the second best ATC has $\left|r\right| = 0.957, \left|\rho\right| = 0.924$.

\textbf{Analysis of the sensitivity of BoP to dataset size.} 
Datasets in DomainNet also vary widely in size (from $48k$ to $173k$) where BoP works well on both dataset-level applications. We further studied the impact of dataset size on DomainNet in dataset testing difficulty by randomly sampling 1\% to 10\% of data from each test set. Correlation of all methods decreases but BoP remains the best: $\left|\rho\right|$ is $ 0.909, 0.857, 0.849$ (decrease from $0.973, 0.914, 0.925$) for BoP, DoC and ATC, respectively.

%-------------------------------

\end{document}